\documentclass[journal]{IEEEtran}
\usepackage{graphicx,subfigure,amsmath,amsfonts,amssymb,algorithm,algorithmic,cite,multirow,bm,
multicol,booktabs,threeparttable,extpfeil}

\begin{document}

\title{FCM-RDpA: TSK Fuzzy Regression Model Construction Using Fuzzy C-Means Clustering, Regularization, DropRule, and Powerball AdaBelief}

\author{Zhenhua~Shi, Dongrui~Wu,
Chenfeng~Guo, Changming~Zhao, Yuqi~Cui, and Fei-Yue~Wang
\thanks{Z.~Shi, D.~Wu, C.~Guo, C.~Zhao and Y.~Cui are with the Ministry of Education Key Laboratory of Image Processing and Intelligent Control, School of Artificial Intelligence and Automation, Huazhong University of Science and Technology, Wuhan, China. Email: \{zhenhuashi, drwu, cfguo, cmzhao, yqcui\}@hust.edu.cn.}
\thanks{F.-Y.~Wang is with the State Key Laboratory for Management and Control of Complex Systems, Institute of Automation, Chinese Academy of Sciences, Beijing, China. Email: feiyue.wang@ia.ac.cn.}
\thanks{D.~Wu and F.-Y. Wang are the corresponding authors.}}

\maketitle

\begin{abstract}
To effectively optimize Takagi-Sugeno-Kang (TSK) fuzzy systems for regression problems, a mini-batch gradient descent with regularization, DropRule, and AdaBound (MBGD-RDA) algorithm was recently proposed. This paper further proposes FCM-RDpA, which improves MBGD-RDA by replacing the grid partition approach in rule initialization by fuzzy c-means clustering, and AdaBound by Powerball AdaBelief, which integrates recently proposed Powerball gradient and AdaBelief to further expedite and stabilize parameter optimization. Extensive experiments on 22 regression datasets with various sizes and dimensionalities validated the superiority of FCM-RDpA over MBGD-RDA, especially when the feature dimensionality is higher. We also propose an additional approach, FCM-RDpAx, that further improves FCM-RDpA by using augmented features in both the antecedents and consequents of the rules.
\end{abstract}


\IEEEpeerreviewmaketitle

\section{Introduction}

Takagi-Sugeno-Kang (TSK) fuzzy models \cite{Takagi1985} have been used in numerous applications \cite{Mendel2017,drwuSeizure2017,drwuIT2FLC2018,drwuTITS2020,drwuMTGA2020}. Their construction consists of three steps: structure selection, parameter initialization, and parameter optimization. This paper proposes a fuzzy c-means clustering \cite{Bezdek1981} plus regularization, DropRule, and Powerball AdaBelief (FCM-RDpA) approach to enhance the recently proposed mini-batch gradient descent with regularization, DropRule, and AdaBound (MBGD-RDA) \cite{Wu2020}, for TSK fuzzy regression model construction.

FCM-RDpA has three enhancements over MBGD-RDA, as shown in Fig.~\ref{fig:1}:

\begin{figure}[htpb] \centering
\includegraphics[width=.9\linewidth,clip]{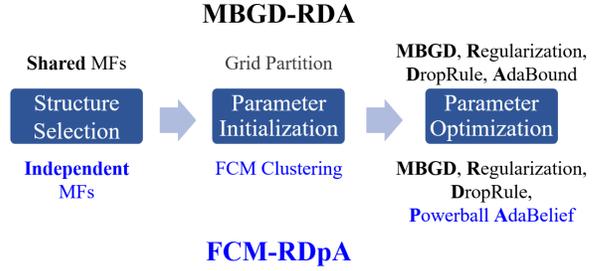}
\caption{A comparison between MBGD-RDA \cite{Wu2020} and our proposed FCM-RDpA in TSK fuzzy regression model construction.} \label{fig:1}
\end{figure}

\begin{enumerate}
\item \emph{Structure selection}: This is mainly related to the structure of the rulebase. There are generally two types of rulebases: i) rulebases with shared membership functions (MFs); and, ii) rulebases with independent MFs.

    Assume there are $M$ features. Then, when each feature has $K$ MFs, a rulebase with shared MFs usually has $K^M$ rules, i.e., the number of rules grows exponentially with the input feature dimensionality. A rulebase with independent MFs can have an arbitrary number of rules, and the MFs in every rule are initialized and optimized independently. Hence, it can better handle high dimensional problems.

    MBGD-RDA \cite{Wu2020} focuses on rulebases with shared MFs, although it can also be used for rulebases with independent MFs with little modification (as used in Section~IV of this paper). FCM-RDpA focuses on rulebases with independent MFs, but it can also be adjusted for the other type of rulebases (as also used in Section~IV of this paper).

\item \emph{Parameter initialization}: Once the rulebase structure is specified, parameter initialization can be performed. MBGD-RDA uses semi-random grid partition initialization \cite{Wu2020} for shared MFs. FCM-RDpA employs fuzzy c-means clustering \cite{Bezdek1981} initialization for independent MFs.

\item \emph{Parameter optimization}: There are mainly four groups of algorithms for TSK fuzzy system parameter optimization \cite{Wu2020}: 1) evolutionary algorithms \cite{Wu2006}; 2) batch gradient descent \cite{Wang1992}; 3) adaptive network-based fuzzy inference system (ANFIS) \cite{Jang1993}, which utilizes batch gradient descent and least squares estimation to optimize the antecedent parameters and consequent parameters, respectively; 4) the recently proposed MBGD-RDA \cite{Wu2020} that utilizes MBGD, regularization, DropRule, and AdaBound to optimize antecedent and consequent parameters simultaneously.

The computational cost of evolutionary algorithms is very high,  especially when the training set is large, and the population size and generation number are large. Batch gradient descent cannot be used for big data. Additionally, MBGD has shown superiority over batch gradient descent in training single-input rule modules fuzzy systems \cite{Matsumura2017} and Mamdani neuro-fuzzy systems \cite{NakasimaLopez2019}. Wu et al. \cite{Wu2020} recently also showed that MBGD-RDA outperformed ANFIS in training TSK fuzzy regression models. So, MBGD is still used in our proposed approach.

There are hundreds of MBGD optimization approaches \cite{Schmidt2020}. Popular representative ones include  stochastic gradient descent \cite{Robbins1951} with momentum (SGDM) \cite{Qian1999}, Adam \cite{Kingma2015}, and AdaBound \cite{Luo2019}. Adam has been used to optimize the parameters of single-input rule modules fuzzy regression models \cite{Matsumura2017}. AdaBound has been used to replace Adam in optimizing the parameters of TSK fuzzy models for regression \cite{Wu2020} and classification\cite{Cui2020a}. However, since the lower and upper bounds of the learning rates in AdaBound are solely determined by the number of mini-batch iterations, AdaBound may lead to local minimum and instability.

To overcome these limitations, our proposed FCM-RDpA enhances MBGD-RDA by replacing AdaBound with Powerball AdaBelief, which combines Powerball gradients \cite{Yuan2019,Zhou2020} and AdaBelief \cite{Zhuang2020} so that the gradients and the learning rates are simultaneously adaptively changed.
\end{enumerate}

In addition to the aforementioned three enhancements, we also propose a variant of FCM-RDpA, FCM-RDpAx, which augments the original features with their trainable linear projections in both the antecedents and the consequents to increase the model capacity.

The main contributions of this paper are:
\begin{enumerate}
\item We propose FCM-RDpA, which enhances MBGD-RDA to handle higher dimensional problems, and to improve and stabilize the generalization performance. To our knowledge, Powerball gradients and AdaBelief have never been integrated before, as we do in RDpA.
\item We conducted extensive experiments to demonstrate the superior performance of FCM-RDpA over ridge regression and the state-of-the-art MBGD-RDA.
\item We also propose an additional approach, FCM-RDpAx, that uses trainable augmented features to further enhance FCM-RDpA.
\end{enumerate}

The remainder of this paper is organized as follows: Section~II introduces the TSK fuzzy regression model and MBGD-RDA. Section~III introduces our proposed FCM-RDpA. Section~IV describes the 22 regression datasets and our experimental results, and proposes FCM-RDpAx. Finally, Section~V draws conclusions.

\section{TSK fuzzy Regression Model and MBGD-RDA}

In this section, we first introduce the TSK fuzzy regression model that utilizes Gaussian MFs \cite{drwuEAAI2019}, and then the state-of-the-art model construction approach, MBGD-RDA \cite{Wu2020}. The key notations are summarized in Table~\ref{tab:notations}.

\begin{table}[htpb]\centering
\caption{Notations used in this paper.} \label{tab:notations}
\begin{tabular}{@{}c|l@{}}
\toprule
Notation                 & \multicolumn{1}{c}{Meaning}                         \\ \midrule
$N$                      & Number of labeled training samples                  \\
$M$                      & Feature dimensionality                              \\
$R$                      & Number of rules                                     \\
$\bm{x}_n=(x_{n,1},$     & \multirow{2}{*}{The $n$th training sample}          \\
$\dots,x_{n,M})^T$       &                                                     \\
$y_n$                    & Groundtruth output corresponding to $x_n$           \\
$X_{r,m}$                & Gaussian MF for the $m$th feature in the $r$th rule \\
$w_{r,0},\dots,w_{r,M}$  & Consequent parameters of the $r$th rule             \\
$y_r(\bm{x}_n)$          & Output of the $r$th rule for $\bm{x}_n$             \\
$\mu_{X_{r,m}}(x_{n,m})$ & Membership grade of $x_{n,m}$ on $X_{r,m}$          \\
$f_r(\bm{x}_n)$          & Firing level of $\bm{x}_n$ on the $r$th rule        \\
$y(\bm{x}_n)$            & Output of the TSK fuzzy system for $\bm{x}_n$       \\
$L$                      & $\ell_{2}$ regularized loss function                \\
$\lambda$                & $\ell_{2}$ regularized coefficient; default $0.05$  \\
$M_m$                    & Number of MFs in each input domain                  \\
$N_{bs}$                 & Mini-batch size; default $64$                       \\
$T$                      & Number of training iterations; default $1,000$      \\
$\alpha$                 & Initial learning rate; default $0.01$               \\
$P$                      & DropRule preservation rate; default $0.5$           \\
$\gamma$                 & Powerball power exponent; default $0.5$             \\ \bottomrule
\end{tabular}
\end{table}

\subsection{The TSK Fuzzy Regression Model}

Assume the input $\bm{x}=(x_1,\dots,x_M)^T \in \mathbb{R}^{M\times 1}$, and the TSK fuzzy system has $R$ rules:
\begin{align}
\begin{split}
\text{Rule}_{r}\text{: } & \text{IF } x_{1} \text{ is } X_{r, 1} \text{ and } \cdots \text{ and } x_{M} \text{ is } X_{r, M}, \\
                         & \text{THEN } y_{r}(\bm{x})=w_{r, 0}+\sum_{m=1}^{M} w_{r, m} x_{m},
\end{split}
\end{align}
where $X_{r,m}$ ($r=1,\dots,R$; $m=1,\dots,M$) are fuzzy sets, and $w_{r,0}$ and $w_{r,m}$ are consequent parameters.

The membership grade of $x_m$ on a Gaussian MF $X_{r,m}$ is:
\begin{align} \label{eq:mu}
\mu_{X_{r, m}}\left(x_{m}\right)=\exp \left(-\frac{\left(x_{m}-c_{r, m}\right)^{2}}{2 \sigma_{r, m}^{2}}\right),
\end{align}
where $c_{r, m}$ is the center of the Gaussian MF, and $\sigma_{r, m}$ the standard deviation.

The firing level of Rule$_r$ is:
\begin{align} \label{eq:fl}
f_{r}(\bm{x})=\prod_{m=1}^{M} \mu_{X_{r, m}}\left(x_{m}\right),
\end{align}
and the output of the TSK fuzzy system is:
\begin{align} \label{eq:y}
y(\bm{x})=\frac{\sum_{r=1}^{R} f_{r}(\bm{x}) y_{r}(\bm{x})}{\sum_{r=1}^{R} f_{r}(\bm{x})}.
\end{align}

\subsection{MBGD-RDA}

Inspired by the connections between TSK fuzzy systems and neural networks \cite{Wu2019}, MBGD-RDA \cite{Wu2020} was recently proposed to integrate MBGD, regularization, DropRule, and AdaBound for effective optimization of TSK fuzzy regression models.

Assume there are $N$ training samples $\{\bm{x}_n,y_n\}_{n=1}^N$, where $\bm{x}_n=(x_{n,1},\dots,x_{n,M})^T \in \mathbb{R}^{M\times 1}$. In each iteration, MBGD-RDA randomly selects $N_{bs}\in [1,N]$ training samples, computes the gradients from them, and then updates the antecedent and consequent parameters of the TSK fuzzy model.

MBGD-RDA uses the following $\ell_{2}$ regularized loss function:
\begin{align} \label{eq:R}
L=\frac{1}{2} \sum_{n=1}^{N_{b s}}\left[y_{n}-y\left(\bm{x}_{n}\right)\right]^{2}+\frac{\lambda}{2} \sum_{r=1}^{R} \sum_{m=1}^{M} w_{r, m}^{2},
\end{align}
where $\lambda \geq 0$ is a regularization parameter. Note that $w_{r, 0}$ ($r=1,\dots,R$) are not regularized in (\ref{eq:R}).

Similar to the idea of Dropout \cite{Srivastava2014} in deep learning to improve generalization, DropRule with preservation rate $P\in (0,1]$ was proposed \cite{Wu2020}. For each training sample in the mini-batch, each rule is chosen with probability $P$ to be used for output prediction, gradient computation, and parameter updating. In testing, all rules are used for prediction, just as in a traditional TSK fuzzy system.

AdaBound \cite{Luo2019} was used for parameter optimization. AdaBound can be viewed as Adam \cite{Kingma2015} with dynamic bounds of learning rates, where the bounds get tighter as the training iterations increase.

\section{FCM-RDpA}

Our proposed FCM-RDpA enhances MBGD-RDA in TSK fuzzy regression model construction. Specifically, it uses independent MFs, initialized by FCM clustering, and Powerball AdaBelief for better generalization. Its pseudo-code is given in Algorithm~\ref{alg:RDpA}. Matlab implementation is available at https://github.com/ZhenhuaShi/FCM-RDpA.

\begin{algorithm}
\caption{The proposed FCM-RDpA algorithm for TSK fuzzy regression model construction. $\odot$ is element-wise product of vectors. By default $\beta_1=0.9$, $\beta_2=0.999$, and $\epsilon=10^{-8}$.} \label{alg:RDpA}
\begin{algorithmic}
\REQUIRE $N$ labeled training samples $\{\bm{x}_n,y_n\}_{n=1}^N$, where $\bm{x}_n=(x_{n,1},\dots,x_{n,M})^T \in \mathbb{R}^{M\times 1}$; \\
$N_v$ labeled validation samples $\{\bm{x}_n,y_n\}_{n=N+1}^{N+N_v}$; \\
$R$, the number of rules; \\
$T$, the maximum number of training iterations; \\
$N_{bs}\in [1,N]$, the mini-batch size; \\
$\lambda$, the $\ell_{2}$ regularized coefficient;\\
$P$, the DropRule preservation rate; \\
$\alpha$, the initial learning rate; \\
$\gamma$, the Powerball power exponent.
\ENSURE The model parameter vector $\boldsymbol{\theta}$.
\STATE \texttt{// FCM clustering initialization}
\STATE Perform FCM clustering ($c=R$) on $\{\bm{x}_n\}_{n=1}^N$;
\STATE Denote the $r$-th cluster center as $\bar{\boldsymbol{c}}_{r}=\left[\bar{c}_{r, 1}, \dots, \bar{c}_{r, M}\right]$, and the corresponding memberships as $\boldsymbol{u}_{r}=\left[u_{r, 1}, \dots, u_{r, N}\right]$;
\FOR{$r=1$ to $R$}
\STATE Initialize $w_{r, 0}=\sum_{n=1}^{N} y_{n} u_{r, n} / \sum_{n=1}^{N} u_{r, n}$;
\FOR{$m=1$ to $M$}
\STATE Initialize $w_{r, m}=0$, $c_{r, m}=\bar{c}_{r, m}$, and $\sigma_{r, m}$ as $\boldsymbol{u}_{r}$ weighted standard deviation of $\{x_{n,m}\}_{n=1}^N$;
\ENDFOR
\ENDFOR
\STATE $\boldsymbol{\theta}_{0}$ is the concatenation of all $c_{r, m}$, $\sigma_{r, m}$, $w_{r, 0}$, and $w_{r, m}$;
\FOR{$t=0$ to $T-1$}
\STATE $\bm{g}_{t}=\frac{\partial L}{\partial \boldsymbol{\theta}_{t}}$;
\STATE Randomly select $N_{bs}$ training samples;
\FOR{$n=1$ to $N_{bs}$}
\FOR{$r=1$ to $R$}
\STATE \texttt{// DropRule}
\STATE $f_r(\bm{x}_n)=0$;
\STATE Generate $p$, a uniformly distributed scalar in $[0,1]$;
\IF{$p\leqslant P$}
\STATE Compute the firing level $f_r(\bm{x}_n)$ by (\ref{eq:mu}) and (\ref{eq:fl});
\ENDIF
\ENDFOR
\STATE Compute the output $y(\bm{x}_n)$ by (\ref{eq:y});
\STATE $\bm{g}_t=\bm{g}_t+\frac{\partial L}{\partial y(\bm{x}_n)}\frac{\partial y(\bm{x}_n)}{\partial \boldsymbol{\theta}_{t}}$;
\ENDFOR
\STATE \texttt{// Powerball AdaBelief}
\STATE $\bm{g}_{t}=\operatorname{sign}(\bm{g}_t) \odot |\bm{g}_t|^{\gamma}$;
\STATE $\bm{m}_{t}=(1-\beta_1)\sum_{i=1}^t \beta_1^{t-i}\bm{g}_i$;
\STATE $\bm{v}_{t}=(1-\beta_2)\sum_{i=1}^t \beta_2^{t-i}(\bm{g}_i-\bm{m}_i)\odot(\bm{g}_i-\bm{m}_i)$;
\STATE $\boldsymbol{\alpha}_{t}=\alpha / (\sqrt{\bm{v}_{t}}+\epsilon)$;
\STATE $\boldsymbol{\theta}_{t+1}=\boldsymbol{\theta}_{t}-\boldsymbol{\alpha}_t\odot \bm{m}_t$;
\STATE Compute the firing levels $\{f_r(\bm{x}_n)\}_{n=N+1}^{N+N_v}$ by (\ref{eq:mu}) and (\ref{eq:fl});
\STATE Compute the outputs $\{y(\bm{x}_n)\}_{n=N+1}^{N+N_v}$ by (\ref{eq:y});
\STATE ${\rm{RMSE}}^{\rm{Val.}}_{t+1}=\sqrt{\frac{1}{N_v} \sum_{n=N+1}^{N+N_v}\left[y_{n}-y\left(\bm{x}_{n}\right)\right]^{2}}$;
\ENDFOR
\STATE $t^*=\arg\min_t\rm{RMSE}^{\rm{Val.}}_t$;
\STATE Return $\boldsymbol{\theta}=\boldsymbol{\theta}_{t^*}$.
\end{algorithmic}
\end{algorithm}

\subsection{FCM Clustering Initialization}

Assume the feature dimensionality is $M$, and each feature has $M_m$ Gaussian MFs. Then, the total number of rules of MBGD-RDA is $R=\prod_{m=1}^{M} M_{m}$, which may be too large for high-dimensional datasets. MBGD-RDA \cite{Wu2020} solved this by performing principal component analysis (PCA) \cite{Jolliffe2002} to reduce the feature dimensionality $M$ to at most 5. However, information may be lost during this process \cite{Shi2020a}, and the final regression performance is hence affected.

\begin{figure}[htpb] \centering
\includegraphics[width=.9\linewidth,clip]{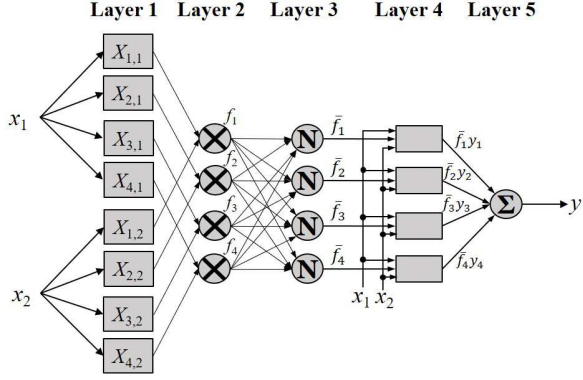}
\caption{Independent MFs to construct TSK fuzzy regression model in our proposed FCM-RDpA. } \label{fig:2}
\end{figure}

To overcome this limitation, our proposed FCM-RDpA enhances MBGD-RDA by replacing the shared MFs with independent ones, and hence to decouple the number of rules from the number of MFs in each input domain, as shown in Fig.~\ref{fig:2}.

Once the model structure is determined, the next step is to initialize the parameters. For shared MFs, grid partition initialization, which generates input MFs by uniformly partitioning the input domains, is frequently used. As a contrast, clustering is usually used to initialize the independent MFs.

More specifically, FCM-RDpA employs FCM clustering \cite{Bezdek1981} to partition the training samples into $R$ fuzzy clusters, each corresponding to one rule, as shown in Algorithm~\ref{alg:RDpA}. When the training set is too large to perform clustering efficiently, we can randomly sample a small subset of it for clustering.

\subsection{Powerball AdaBelief Optimization}

For parameter optimization, compared with MBGD-RDA, FCM-RDpA also integrates MBGD, regularization and DropRule for gradient computation, but replaces AdaBound with Powerball AdaBelief for parameter updating, as shown in Algorithm~\ref{alg:RDpA}.

Specifically, Powerball stochastic gradient descent \cite{Yuan2019,Zhou2020} adapts the gradients by directly adding a Powerball power exponent $\gamma\in [0,1)$ so that gradient explosion caused by outliers can be alleviated\footnote{TSK fuzzy regression models are shallow, compared with deep neural networks. Thus, there is no need to consider gradient vanishing \cite{Zhou2020} here.}.

Moreover, unlike AdaBound \cite{Luo2019} that determines the lower and upper bounds of the learning rates solely by the number of mini-batch iterations, AdaBelief \cite{Zhuang2020} takes into consideration the curvature information of the loss function and adapts the learning rate by the belief in observed gradients. Detailed comparisons of AdaBound and AdaBelief are shown in Table~\ref{tab:GD} in the next section.

Including AdaBound and Powerball AdaBelief, there are hundreds of MBGD optimization approaches \cite{Schmidt2020}. A generic framework is shown in Algorithm~\ref{alg:MBGD} to encapsulate most of them, and Powerball AdaBelief is used as a specific example. Note that when the Powerball power exponent $\gamma=1$, different definitions of $\{\boldsymbol{\phi}_{t},\boldsymbol{\psi}_{t},\boldsymbol{\omega}_t\}_{t=1}^T$ specified in Table~\ref{tab:GD} lead to different vanilla MBGD approaches. When the Powerball power exponent $\gamma\in [0,1)$, different definitions of $\{\boldsymbol{\phi}_{t},\boldsymbol{\psi}_{t},\boldsymbol{\omega}_t\}_{t=1}^T$ lead to different powered MBGD approaches.

\begin{algorithm}
\caption{Generic framework\protect\footnote{} of MBGD optimization approaches. Powerball AdaBelief is used as a specific example. All vector products and powers are element-wise.} \label{alg:MBGD}
\begin{algorithmic}
\STATE $\boldsymbol{\theta}_{t}$: Model parameter vector in the $t$th training iteration;
\STATE $\mathcal{F}$: Feasible set with bounded diameter;
\STATE $\Pi_{\mathcal{F},\sqrt{\bm{v}_{t}}}(\hat{\boldsymbol{\theta}}_{t+1})$: Projection of $\hat{\boldsymbol{\theta}}_{t+1}$ onto the feasible set $\mathcal{F}$,
\STATE \hspace*{2mm} i.e., $\Pi_{\mathcal{F},\sqrt{\bm{v}_{t}}}(\hat{\boldsymbol{\theta}}_{t+1})=\operatorname{argmin}_{\boldsymbol{\theta} \in \mathcal{F}}\left\|\sqrt[4]{\bm{v}_{t}}(\boldsymbol{\theta}-\hat{\boldsymbol{\theta}}_{t+1})\right\|$.
\REQUIRE Initial model parameter vector $\boldsymbol{\theta}_{0} \in \mathcal{F}$;\\
Initial learning rate $\alpha$;\\
Powerball power exponent $\gamma$;\\
Sequence of functions $\{\boldsymbol{\phi}_{t},\boldsymbol{\psi}_{t},\boldsymbol{\omega}_t\}_{t=1}^T$ (specified in Table~\ref{tab:GD}).
\ENSURE The final $\boldsymbol{\theta}_{T}$.
\FOR{$t=0$ to $T-1$}
\STATE $\bm{g}_{t}=\operatorname{sign}\left(\nabla_{\boldsymbol{\theta}_{t}} L\right)|\nabla_{\boldsymbol{\theta}_{t}} L|^{\gamma}$
\STATE $\bm{m}_{t}=\boldsymbol{\phi}_{t}(\bm{g}_1,...,\bm{g}_{t})\xlongequal{\rm{e.g.}} (1-\beta_1)\sum_{i=1}^t \beta_1^{t-i}\bm{g}_i$
\STATE $\bm{v}_{t}=\boldsymbol{\psi}_{t}(\bm{g}_1,...,\bm{g}_{t})\xlongequal{\rm{e.g.}} (1-\beta_2)\sum_{i=1}^t \beta_2^{t-i}(\bm{g}_i-\bm{m}_i)^2$
\STATE $\boldsymbol{\alpha}_{t}=\boldsymbol{\omega}_t(\alpha,t,\bm{v}_t)\xlongequal{\rm{e.g.}} \alpha / (\sqrt{\bm{v}_{t}}+\epsilon)$
\STATE $\hat{\boldsymbol{\theta}}_{t+1}=\boldsymbol{\theta}_{t}-\boldsymbol{\alpha}_t \bm{m}_t$
\STATE $\boldsymbol{\theta}_{t+1}=\Pi_{\mathcal{F},\sqrt{\bm{v}_{t}}}(\hat{\boldsymbol{\theta}}_{t+1})$
\ENDFOR
\end{algorithmic}
\end{algorithm}
\footnotetext{Similar frameworks were shown in \cite{Reddi2018,Luo2019}. Our version is more general.}

\section{Experiments}

This section validates the performance of our proposed FCM-RDpA on multiple datasets with various sample numbers and feature dimensionalities.

\subsection{Datasets}

Twenty-two regression datasets from the CMU StatLib Datasets Archive\footnote{http://lib.stat.cmu.edu/datasets/} and the UCI Machine Learning Repository\footnote{http://archive.ics.uci.edu/ml/index.php} were used in our experiments. Their summary is shown in Table~\ref{tab:data}.

\begin{table}[htpb] \centering \setlength{\tabcolsep}{0.8mm}
\caption{Summary of the 22 regression datasets.} \label{tab:data}
\begin{threeparttable}
\begin{tabular}{@{}cccccc@{}}
\toprule
\multirow{2}{*}{Index} & \multirow{2}{*}{Dataset}          & \multirow{2}{*}{Source} & No. of   & No. of   & Feature               \\
                       &                                   &                         & Samples & Features & Type                  \\ \midrule
1                      & Concrete-CS\tnote{1}              & UCI                     & 103      & 7        & Numerical             \\
2                      & Concrete-Flow\tnote{1}            & UCI                     & 103      & 7        & Numerical             \\
3                      & Concrete-Slump\tnote{1}           & UCI                     & 103      & 7        & Numerical             \\
4                      & Tecator-fat\tnote{2}              & StatLib                 & 240      & 22       & Numerical             \\
5                      & Tecator-moisture\tnote{2}         & StatLib                 & 240      & 22       & Numerical             \\
6                      & Tecator-protein\tnote{2}          & StatLib                 & 240      & 22       & Numerical             \\
7                      & Yacht\tnote{3}                    & UCI                     & 308      & 6        & Numerical             \\
8                      & autoMPG\tnote{4}                  & UCI                     & 392      & 9        & Categorical, Numerical \\
9                      & NO2\tnote{5}                      & StatLib                 & 500      & 7        & Numerical             \\
10                     & PM10\tnote{6}                     & StatLib                 & 500      & 7        & Numerical             \\
11                     & Housing\tnote{7}                  & UCI                     & 506      & 13       & Numerical             \\
12                     & CPS\tnote{8}                      & StatLib                 & 534      & 19       & Categorical, Numerical \\
13                     & Efficiency-Cool\tnote{9}          & UCI                     & 768      & 8        & Numerical             \\
14                     & Efficiency-Heat\tnote{9}          & UCI                     & 768      & 8        & Numerical             \\
15                     & Concrete\tnote{10}                 & UCI                     & 1,030    & 8        & Numerical             \\
16                     & Airfoil\tnote{11}                  & UCI                     & 1,503    & 5        & Numerical             \\
17                     & Wine-red\tnote{12}                 & UCI                     & 1,599    & 11       & Numerical             \\
18                     & Abalone\tnote{13}                  & UCI                     & 4,177    & 7        & Numerical             \\
19                     & Abalone-OH\tnote{13}               & UCI                     & 4,177    & 10       & Categorical, Numerical \\
20                     & Wine-white\tnote{12}               & UCI                     & 4,898    & 11       & Numerical             \\
21                     & PowerPlant\tnote{14}               & UCI                     & 9,568    & 4        & Numerical             \\
22                     & Protein\tnote{15}                  & UCI                     & 45,730   & 9        & Numerical             \\ \bottomrule
\end{tabular}
\begin{tablenotes}
\item[1] https://archive.ics.uci.edu/ml/datasets/Concrete+Slump+Test
\item[2] http://lib.stat.cmu.edu/datasets/tecator
\item[3] http://archive.ics.uci.edu/ml/datasets/Yacht+Hydrodynamics
\item[4] https://archive.ics.uci.edu/ml/datasets/Auto+MPG
\item[5] http://lib.stat.cmu.edu/datasets/NO2.dat
\item[6] http://lib.stat.cmu.edu/datasets/PM10.dat
\item[7] https://archive.ics.uci.edu/ml/machine-learning-databases/housing/
\item[8] http://lib.stat.cmu.edu/datasets/CPS\_85\_Wages
\item[9] http://archive.ics.uci.edu/ml/datasets/Energy+efficiency
\item[10] http://archive.ics.uci.edu/ml/datasets/Concrete+Compressive+Strength
\item[11] https://archive.ics.uci.edu/ml/datasets/Airfoil+Self-Noise
\item[12] https://archive.ics.uci.edu/ml/datasets/Wine+Quality
\item[13] https://archive.ics.uci.edu/ml/datasets/Abalone
\item[14] https://archive.ics.uci.edu/ml/datasets/Combined+Cycle+Power+Plant
\item[15] https://archive.ics.uci.edu/ml/datasets/Physicochemical+Properties+of+ \\
Protein+Tertiary+Structure
\end{tablenotes}
\end{threeparttable}
\end{table}

All categorical features were encoded by one-hot coding. All features were then $z$-normalized to have zero mean and unit variance, and the output mean was also subtracted.

\subsection{Algorithms}

We compare the performances of the following seven algorithms:
\begin{enumerate}
\item \emph{Ridge regression (RR)}: The inputs are the original (unreduced) features. Ridge coefficient $\lambda=0.05$ was used. This is our baseline for comparison.
\item \emph{PCA-GP-RDA}: It uses PCA to reduce the number of features to $\log_{M_m} (R)$ ($M_m=2$ is used in this paper), grid partition to initialize the shared Gaussian MFs, and then MBGD-RDA for optimization.
\item \emph{PCA-GP-RDpA}: It is identical to PCA-GP-RDA, except that RDpA replaces RDA in parameter optimization.
\item \emph{PCA-FCM-RDA}: It uses PCA to reduce the number of features to $\log_{M_m} (R)$ ($M_m=2$ is used in this paper), FCM clustering to initialize the independent Gaussian MFs, and then MBGD-RDA for optimization.
\item \emph{PCA-FCM-RDpA}: It is identical to PCA-FCM-RDA, except that RDpA replaces RDA in parameter optimization.
\item \emph{FCM-RDA}: It uses the original features (without PCA dimensionality reduction), FCM clustering to initialize the independent Gaussian MFs, and then MBGD-RDA for optimization.
\item \emph{FCM-RDpA}.
\end{enumerate}
A comparison of the latter six algorithms are shown in Table~\ref{tab:AA}. The maximum number of iterations was $1,000$. Following \cite{Wu2020}, default parameter values were: initial learning rate $\alpha=0.01$, DropRule preservation rate $P=0.5$, $\ell_{2}$ regularization coefficient $\lambda=0.05$, and mini-batch size $N_{bs}=64$. Additionally, Powerball power exponent $\gamma=0.5$.

\begin{table}[htpb] \centering
\caption{Comparison of the six algorithms.} \label{tab:AA}
\begin{tabular}{@{}l|ccccc@{}}
\toprule
\multirow{2}{*}{Algorithm} & \multirow{2}{*}{PCA} & Shared       & Independent  & MBGD         & MBGD         \\
                           &                      & MFs (GP)     & MFs (FCM)    & -RDA         & -RDpA        \\ \midrule
PCA-GP-RDA                 & $\checkmark$         & $\checkmark$ &              & $\checkmark$ &              \\
PCA-GP-RDpA                & $\checkmark$         & $\checkmark$ &              &              & $\checkmark$ \\ \midrule
PCA-FCM-RDA                & $\checkmark$         &              & $\checkmark$ & $\checkmark$ &              \\
PCA-FCM-RDpA               & $\checkmark$         &              & $\checkmark$ &              & $\checkmark$ \\ \midrule
FCM-RDA                    &                      &              & $\checkmark$ & $\checkmark$ &              \\
FCM-RDpA                   &                      &              & $\checkmark$ &              & $\checkmark$ \\ \bottomrule
\end{tabular}
\end{table}

Note that when the number of features was lower than $\log_2 (R)$, e.g., when $R=32$ and $R=64$ on PowerPlant with 4 features, and when $R=64$ on Airfoil with 5 features, PCA cannot be performed. Thus, all algorithms were not evaluated for these cases.

\subsection{Experimental Settings}

We randomly partitioned each dataset into three subsets: 70\% for training, 15\% for validation, and the remaining 15\% for test. This process was repeated eight times and the average root mean squared errors (RMSEs) on the test sets were computed as the performance measure. In each run, we identified the iteration number that gave the smallest validation RMSE, and recorded the corresponding test RMSE. The platform was a laptop computer with AMD Ryzen 5 CPU (3550H@2.10GHz) and 16GB RAM, running Windows 10 x64 and Matlab 2020a.

\subsection{Experiment Results}

\begin{figure*}[htpb] \centering
\subfigure[]{\label{fig:ALGS} \includegraphics[width=\linewidth,clip]{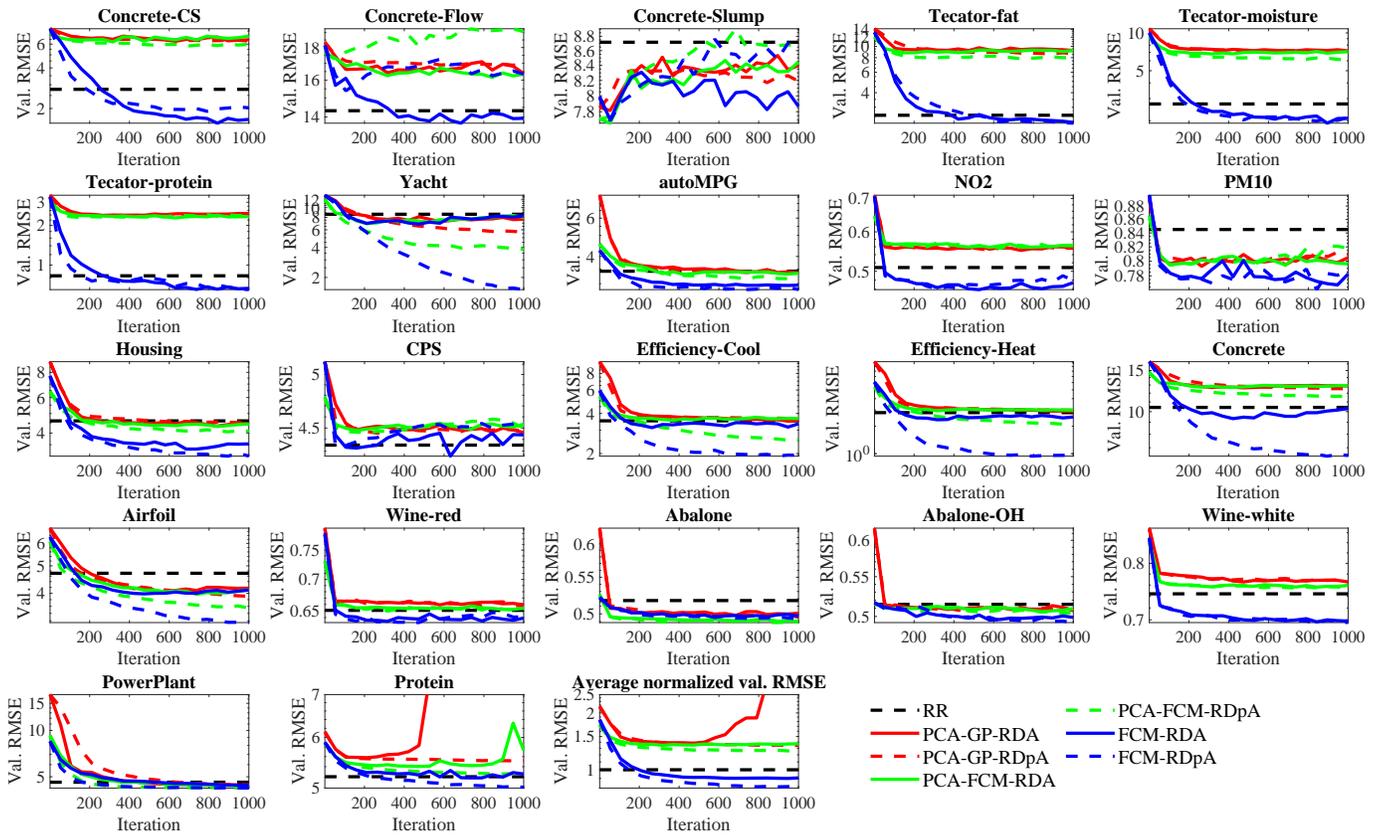}}
\subfigure[]{\label{fig:Rs} \includegraphics[width=\linewidth,clip]{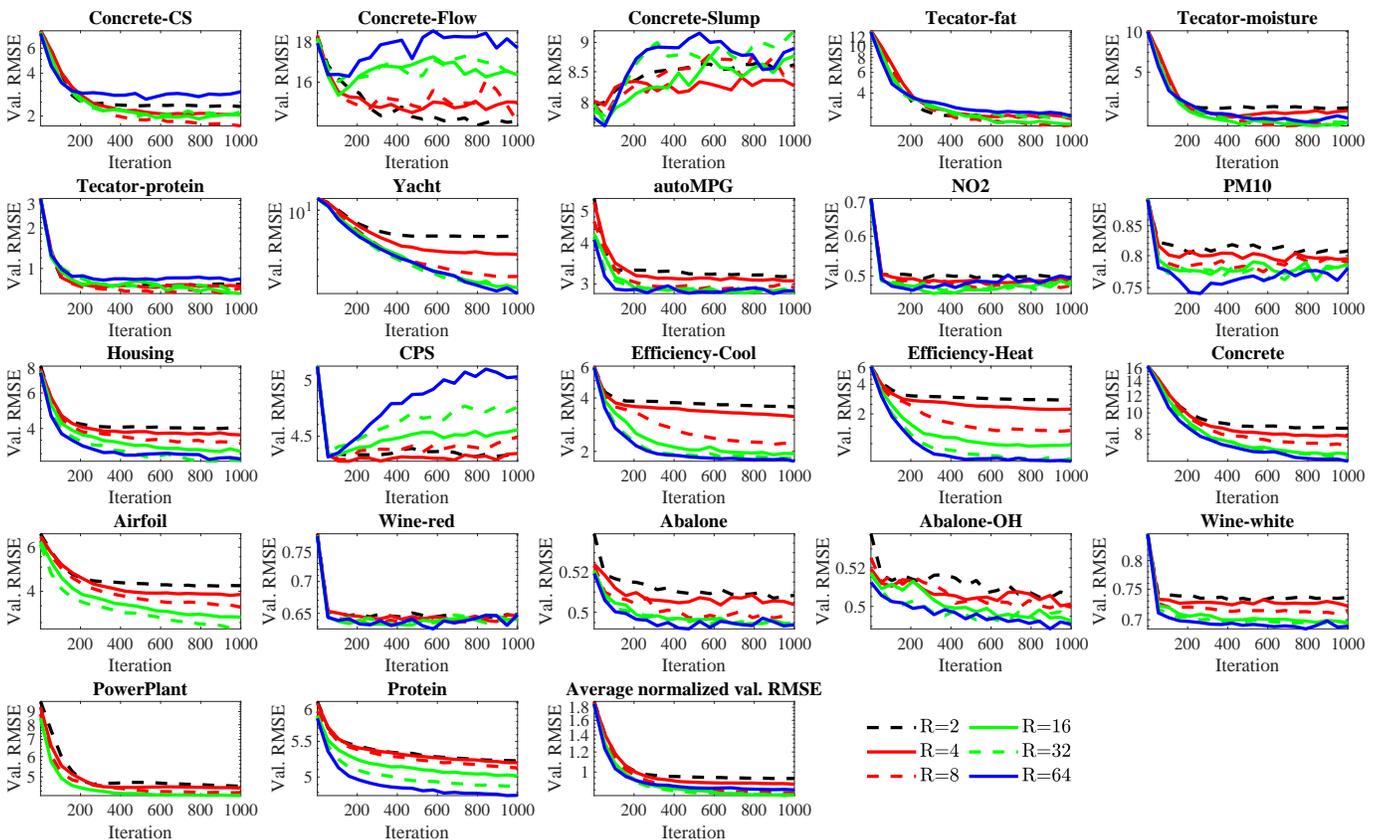}}
\caption{Validation RMSEs of (a) the seven algorithms with $R=16$, and (b) the proposed FCM-RDpA with six different $R$, versus the number of iterations, averaged over eight runs on each dataset. In each subfigure, the last panel shows the average normalized validation RMSEs across the 22 datasets.} \label{fig:3}
\end{figure*}

The validation RMSEs of the seven algorithms ($R=16$) on the 22 datasets, versus the number of training iterations, are shown in Fig.~\ref{fig:ALGS}. Our proposed FCM-RDpA almost always achieved the best validation performance, when the number of iterations was large.

To see the forest for the trees, the last panel of Fig.~\ref{fig:ALGS} also shows the average \emph{normalized} RMSEs of different algorithms. Because the magnitudes of RMSEs from different datasets were dramatically different, it is not meaningful to directly average them across datasets. So, for each dataset, we use the RMSE of RR as the baseline, and normalize the RMSE of each MBGD-based algorithm w.r.t. to it. Finally, we compute the average of the normalized RMSEs across the 22 datasets. The last panel of Fig.~\ref{fig:ALGS} shows that on average our proposed FCM-RDpA achieved the smallest RMSE, given enough training iterations.

The RMSEs of FCM-RDpA w.r.t. different number of training iterations and different number of rules ($R$) are shown in Fig.~\ref{fig:Rs}. The last panel also shows the average normalized RMSEs w.r.t. to the RMSE of RR. Observe that small datasets prefer a smaller number of rules, whereas large datasets prefer a larger number of rules, which is intuitive. The best overall performance was obtained when $R=16$.

Fig.~\ref{fig:4a} shows the average normalized test RMSEs w.r.t. different $R$. Note that the test RMSE of RR did not change with $R$, because it always used all original features. The following observations can be made:

\begin{figure}[htpb] \centering
\subfigure[]{\label{fig:4a}\includegraphics[width=.9\linewidth,clip]{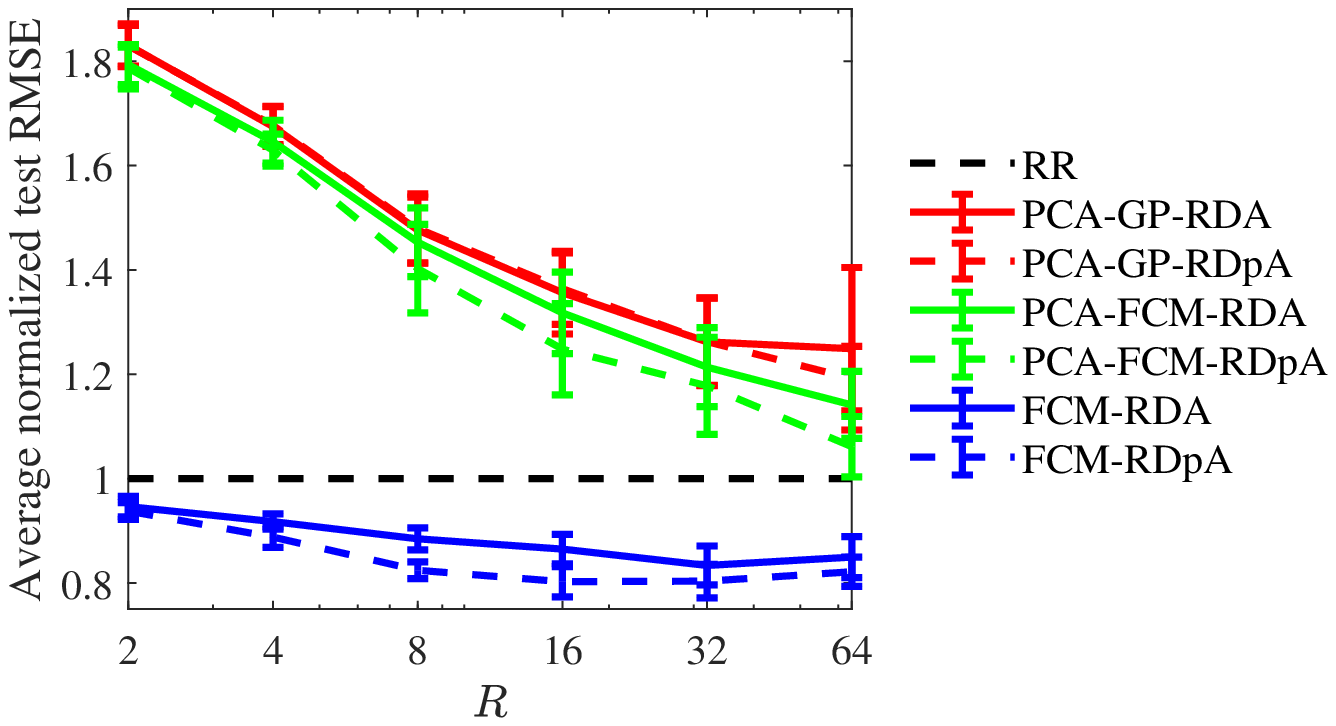}}
\subfigure[]{\label{fig:4b}\includegraphics[width=.9\linewidth,clip]{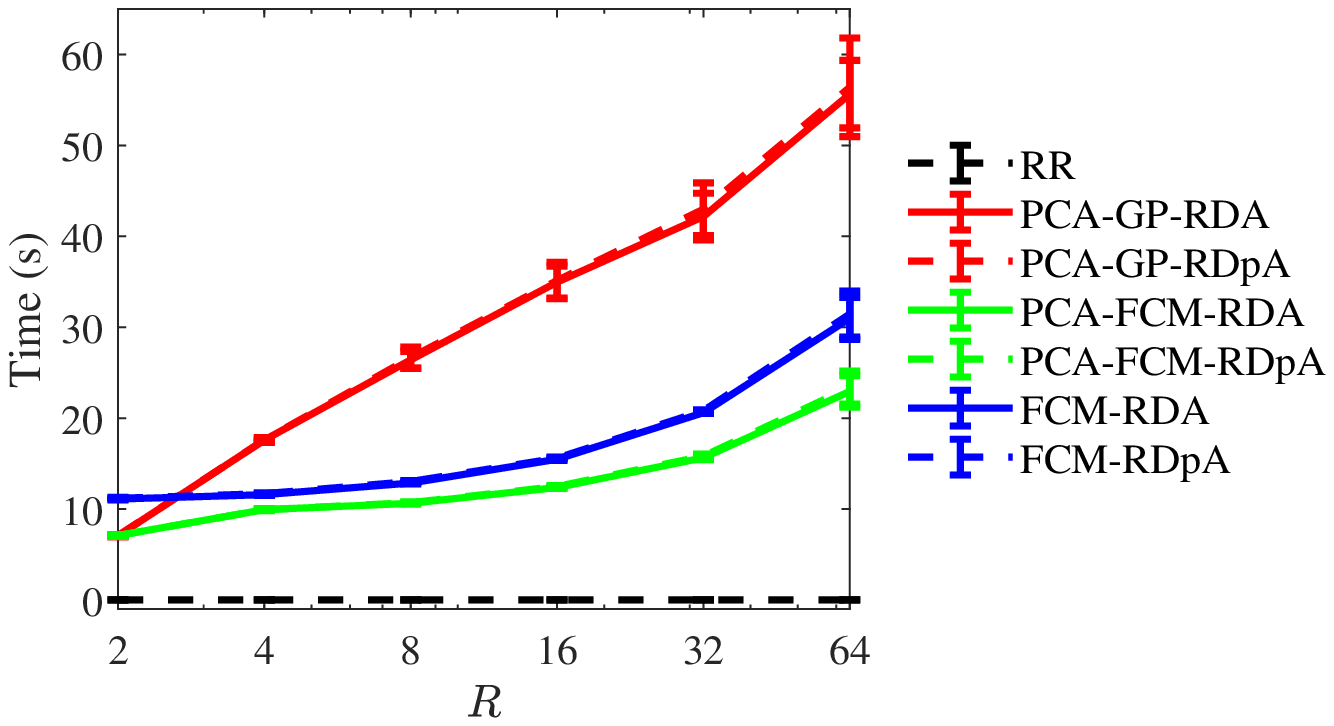}}
\caption{(a) Average normalized test RMSEs and (b) average computing time (seconds; including training, validation and test time) w.r.t. the number of rules $R$, averaged over the 22 regression datasets in eight runs with different data partitions. } \label{fig:4}
\end{figure}

\begin{enumerate}
\item \emph{The effect of PCA}: Comparing PCA-FCM-RDA with FCM-RDA (or PCA-FCM-RDpA with FCM-RDpA), it is clear that PCA resulted in worse performance, due to the reduced dimensionality and loss of information \cite{Shi2020a}. Moreover, the PCA versions always had worse performance than RR, which does not agree with the results in \cite{Wu2020}. This is because in \cite{Wu2020} RR also used the PCA features, whereas here RR used the original (unreduced) features. On the other hand, without PCA, both FCM-RDA and FCM-RDpA outperformed RR. These results suggest that the original (high-dimensional) features should be preferred in TSK fuzzy regression models. Of course, the corresponding training algorithm should be able to effectively deal with potentially high dimensional features.

\item \emph{The effect of shared/independent MFs}: Comparing PCA-GP-RDA with PCA-FCM-RDA (or PCA-GP-RDpA with PCA-FCM-RDpA), it is clear that independent MFs, and the associated FCM initialization, resulted in better performance. This is intuitive, as a TSK fuzzy model using independent MFs has more parameters than one using shared MFs, and hence the former has a larger model capacity.

\item \emph{The effect of optimization approaches}: Comparing PCA-FCM-RDA with PCA-FCM-RDpA (or FCM-RDA with FCM-RDpA), it is clear that RDpA resulted in better regression performance, i.e., the Powerball AdaBelief is more effective than AdaBound. Interestingly, PCA-GP-RDA and PCA-GP-RDpA has similar performances. This may be due to the fact that both models had a very small number of parameters, which are easy to optimize, and hence a sophisticated optimization approach like RDpA may not demonstrate its superiority.

\item \emph{The effect of the number of rules}: Generally, as $R$, the number of rules, increased, the RMSEs decreased for all TSK fuzzy systems. However, RDpA based optimization resulted in fastest convergence, e.g., FCM-RDpA converged at $R=8$.

\item On average, FCM-RDpA achieved the best performance, suggesting that our proposed approach, which integrates independent MFs, FCM clustering initialization, and RDpA optimization, is effective.
\end{enumerate}

The average computing time of the seven algorithms is shown in Fig.~\ref{fig:4b}. RR is much faster, as it has a closed-form solution. PCA-GP-RDA and PCA-GP-RDpA are generally the slowest, because GP initialization requires extra time to enumerate all $R=\prod_{m=1}^{M} M_{m}$ possible combinations of shared MFs for firing levels computation. The computational cost of MBGD-RDA and FCM-RDpA was always almost identical.

\subsection{Parameter Sensitivity of FCM-RDpA} \label{sect:PS}

\begin{figure}[htpb] \centering
\subfigure[]{\label{fig:RMSEvsAlpha}     \includegraphics[width=.85\linewidth,clip]{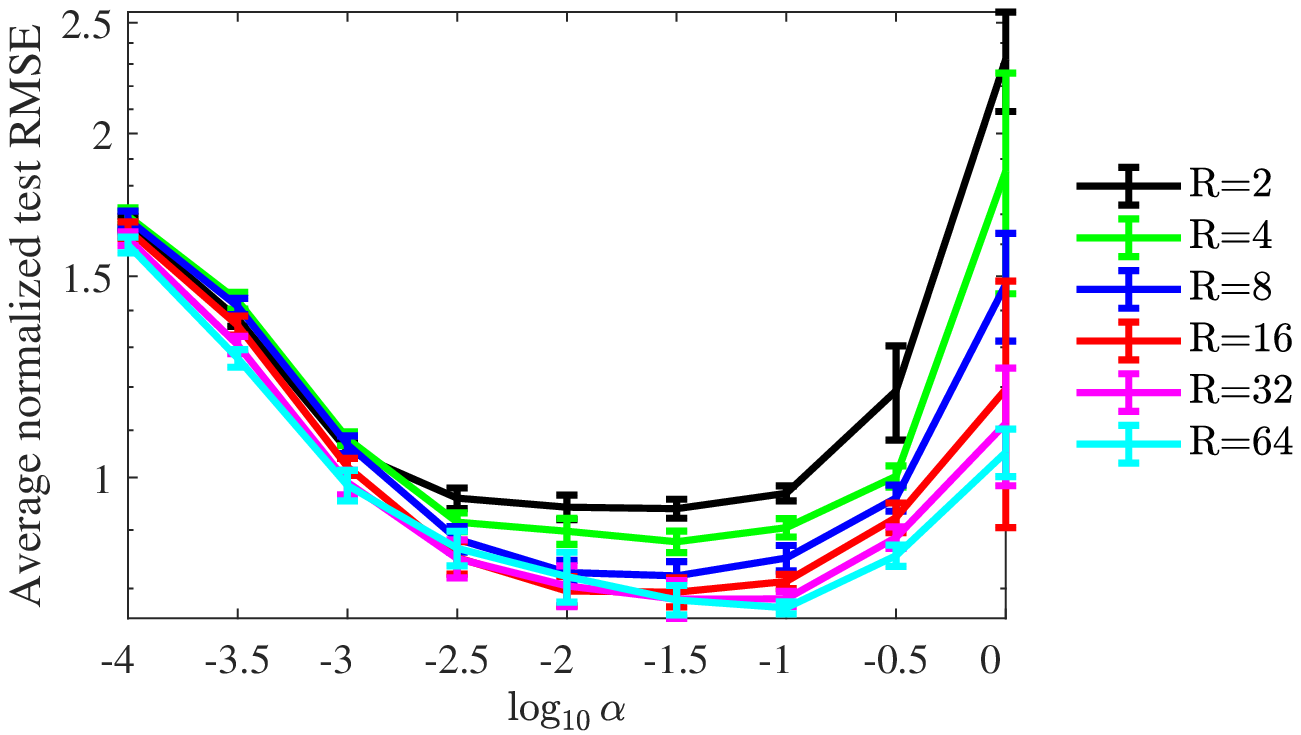}}\hfill
\subfigure[]{\label{fig:RMSEvsP}     \includegraphics[width=.85\linewidth,clip]{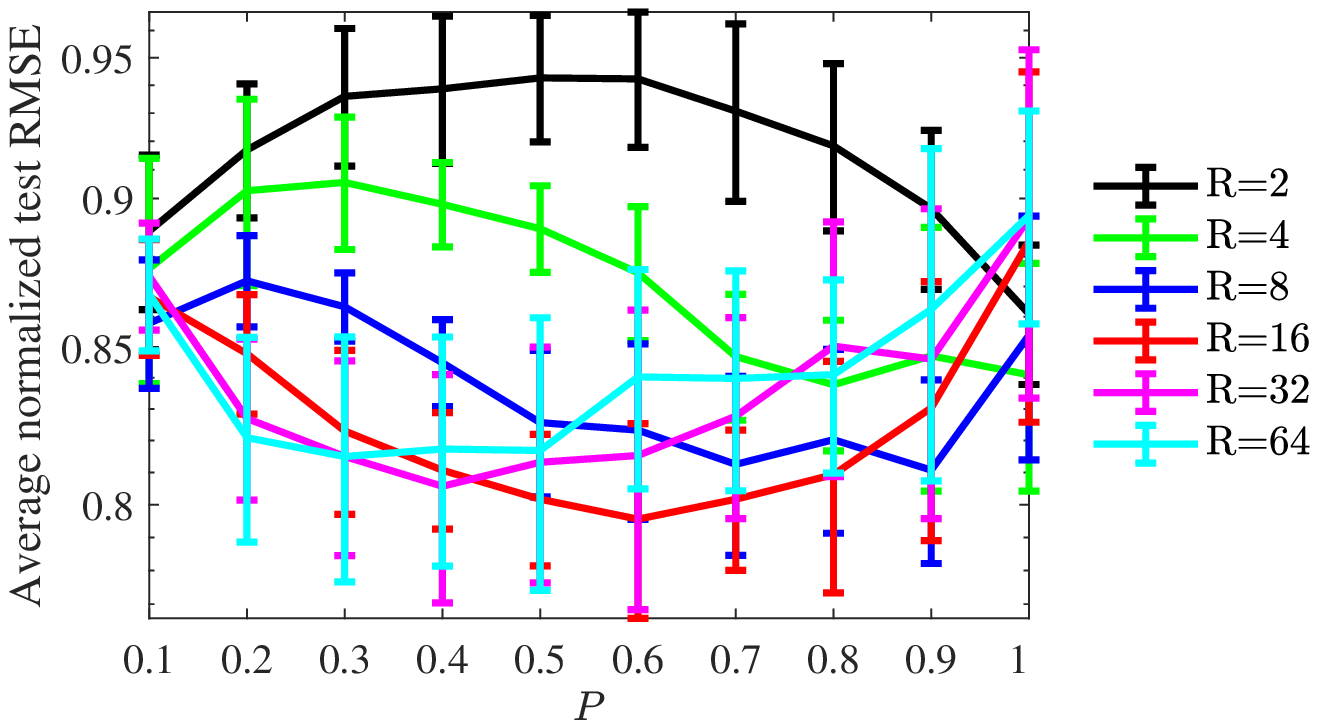}}\hfill
\subfigure[]{\label{fig:RMSEvsGamma}     \includegraphics[width=.85\linewidth,clip]{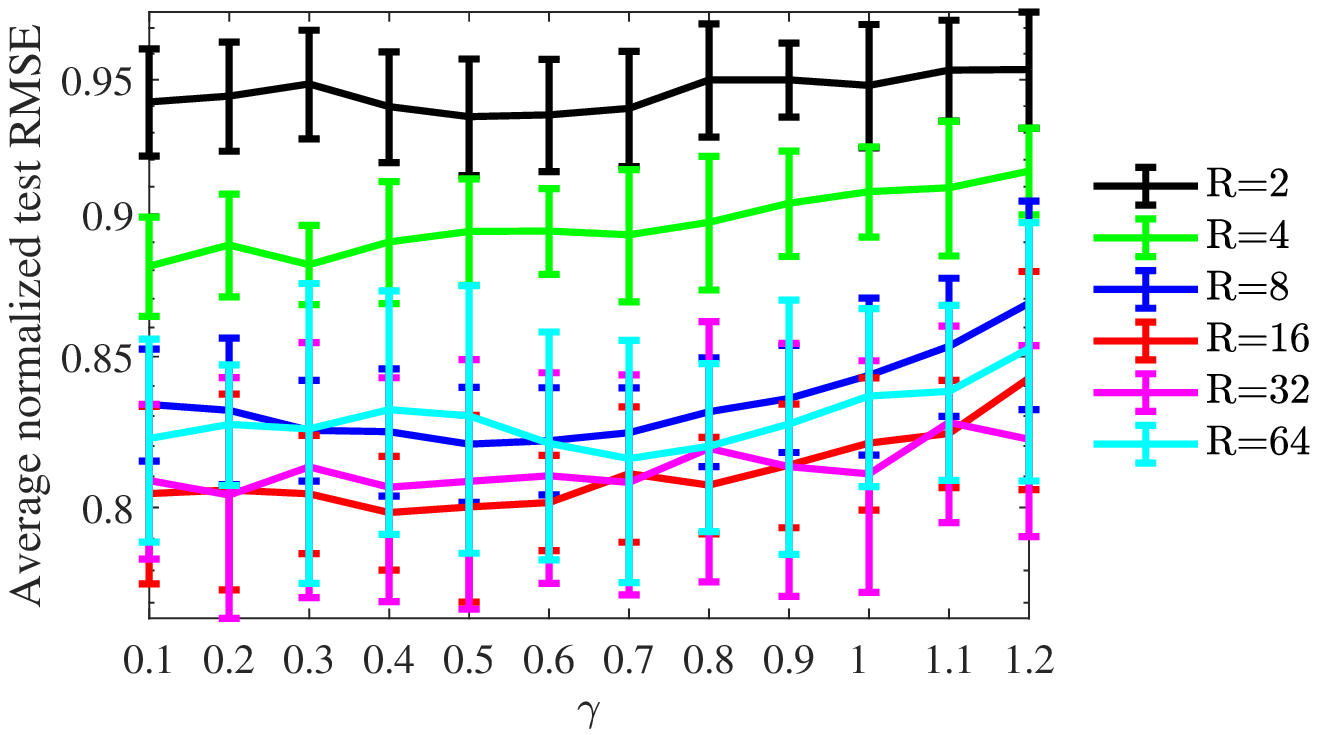}}
\caption{Average normalized test RMSEs of FCM-RDpA versus its three parameters, averaged over the 22 regression datasets in eight runs with different data partitions. (a) Initial learning rate $\alpha$; (b) DropRule preservation rate $P$; (c) Powerball power exponent $\gamma$. When changing the x-axis parameter, the other two parameters were set as default ($\alpha=0.01, P=0.5, \gamma=0.5$). } \label{fig:5}
\end{figure}

The parameter sensitivity of FCM-RDpA is studied next. The average normalized test RMSEs of FCM-RDpA versus its three parameters, averaged over the 22 regression datasets in eight runs with different data partitions, are shown in Fig.~\ref{fig:5}. In general, we conclude that it is safe to choose the learning rate $\alpha\in[0.01,0.1]$, the DropRule preservation rate $P\in[0.5,0.7]$ and the Powerball power exponent $\gamma\in[0.3,0.6]$.

\subsection{Effect of FCM Clustering Initialization}

\begin{figure}[htpb] \centering
\includegraphics[width=.9\linewidth,clip]{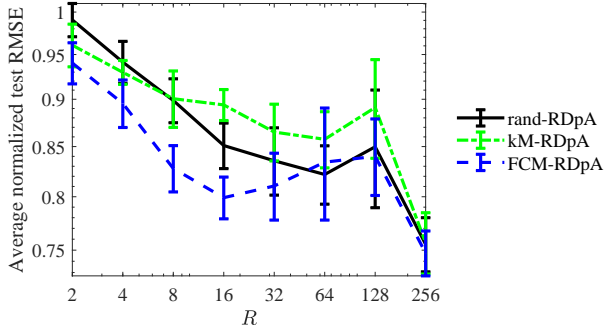}
\caption{Average normalized test RMSEs of rand-RDpA (random initialization), kM-RDpA ($k$-means clustering initialization), and FCM-RDpA (FCM clustering initialization) versus the number of rules $R$, over the 22 datasets. Parameters were set as default (initial learning rate $\alpha=0.01$, DropRule preservation rate $P=0.5$, Powerball power exponent $\gamma=0.5$). } \label{fig:7}
\end{figure}

This section studies the effect of FCM clustering initialization. The average normalized test RMSEs of rand-RDpA (random initialization, which uniformly samples Gaussian MF centers $c_{r, m}$ ($r=1,\dots,R$; $m=1,\dots,M$) and consequent parameters $w_{r, m}$ in $[0,1]$, and Gaussian MF standard deviations $\sigma_{r, m}$ in $[0,5]$), kM-RDpA ($k$-means clustering initialization), and FCM-RDpA (FCM clustering initialization) versus the number of rules $R$ are shown in Fig.~\ref{fig:7}.

FCM clustering initialization resulted in much better performance when the number of rules $R$ is not too large. The three initialization approaches had similar performance when $R$ is very large. So, FCM clustering initialization is preferred.

\subsection{Effect of Powerball AdaBelief Optimization}

For the TSK fuzzy regression model initialized with FCM clustering, this section compares Powerball AdaBelief with seven other MBGD optimization approaches, as shown in Table~\ref{tab:GD}.

\begin{table}[htpb]\centering
\caption{Comparison of eight algorithms with different MBGD optimization approaches.} \label{tab:GD}
\begin{threeparttable}
\begin{tabular}{@{}llc@{}}
\toprule
\multirow{2}{*}{Algorithm}    & \multirow{2}{*}{Vanilla MBGD optimization approach\tnote{1}}                                                                     & Powerball                     \\
                              &                                                                                                                                  & Gradient                      \\ \midrule
\multirow{2}{*}{RDA}          & AdaBound:                                                                                                                        & \multirow{2}{*}{--}           \\
                              & $\boldsymbol{\phi}_t=(1-\beta_1)\sum_{i=1}^t \beta_1^{t-i}\bm{g}_i$                                                              &                               \\
\multirow{2}{*}{RD-pAdaBound} & $\boldsymbol{\psi}_t=(1-\beta_2)\sum_{i=1}^t \beta_2^{t-i}\bm{g}_i^2$                                                            & \multirow{2}{*}{$\checkmark$} \\
                              & $\boldsymbol{\omega}_t=\operatorname{Clip}\left(\alpha / (\sqrt{\boldsymbol{\psi}_t}+\epsilon), \eta_{l}(t), \eta_{u}(t)\right)$ &                               \\ \midrule
\multirow{2}{*}{RD-SGDM}      & SGDM:                                                                                                                            & \multirow{2}{*}{--}           \\
                              & $\boldsymbol{\phi}_t=\sum_{i=1}^t \beta_1^{t-i}\bm{g}_i$                                                                         &                               \\
\multirow{2}{*}{RD-pSGDM}     & $\boldsymbol{\psi}_t=\bm{1}$                                                                                                     & \multirow{2}{*}{$\checkmark$} \\
                              & $\boldsymbol{\omega}_t=\alpha$                                                                                                   &                               \\ \midrule
\multirow{2}{*}{RD-Adam}      & Adam:                                                                                                                            & \multirow{2}{*}{--}           \\
                              & $\boldsymbol{\phi}_t=(1-\beta_1)\sum_{i=1}^t \beta_1^{t-i}\bm{g}_i$                                                              &                               \\
\multirow{2}{*}{RD-pAdam}     & $\boldsymbol{\psi}_t=(1-\beta_2)\sum_{i=1}^t \beta_2^{t-i}\bm{g}_i^2$                                                            & \multirow{2}{*}{$\checkmark$} \\
                              & $\boldsymbol{\omega}_t=\alpha / (\sqrt{\boldsymbol{\psi}_{t}}+\epsilon)$                                                         &                               \\ \midrule
\multirow{2}{*}{RD-AdaBelief} & AdaBelief:                                                                                                                       & \multirow{2}{*}{--}           \\
                              & $\boldsymbol{\phi}_t=(1-\beta_1)\sum_{i=1}^t \beta_1^{t-i}\bm{g}_i$                                                              &                               \\
\multirow{2}{*}{RDpA}         & $\boldsymbol{\psi}_t=(1-\beta_2)\sum_{i=1}^t \beta_2^{t-i}(\bm{g}_i-\boldsymbol{\phi}_{i})^2$                                    & \multirow{2}{*}{$\checkmark$} \\
                              & $\boldsymbol{\omega}_t=\alpha / (\sqrt{\boldsymbol{\psi}_{t}}+\epsilon)$                                                         &                               \\ \bottomrule
\end{tabular}
\begin{tablenotes}
\item[1] All vector powers are element-wise. For simplicity, the debiasing steps are unshown. The usage of $\{\boldsymbol{\phi}_{t},\boldsymbol{\psi}_{t},\boldsymbol{\omega}_t\}$ is shown in Algorithm~\ref{alg:MBGD}. By default, $\beta_1=0.9$, $\beta_2=0.999$, $\epsilon=10^{-8}$, $\eta_{l}(t)=\frac{\alpha(1-\beta_2)t}{(1-\beta_2)t+1}$, and $\eta_{u}(t)=\frac{\alpha\left((1-\beta_2)t+1\right)}{(1-\beta_2)t}$.
\end{tablenotes}
\end{threeparttable}
\end{table}

The average normalized test RMSEs of the eight algorithms versus the number of rules $R$, averaged on the 22 datasets in eight runs with different data partitions, are shown in Fig.~\ref{fig:6}. Since MBGD optimization approaches are sensitive to the initial learning rate $\alpha$, we tuned $\alpha$ from $\{1,0.1,0.01,0.001,0.0001\}$ using the validation RMSE. The other two parameters were set as default (DropRule preservation rate $P=0.5$, Powerball power exponent $\gamma=0.5$). Observe that:

\begin{figure}[htpb] \centering
\includegraphics[width=.9\linewidth,clip]{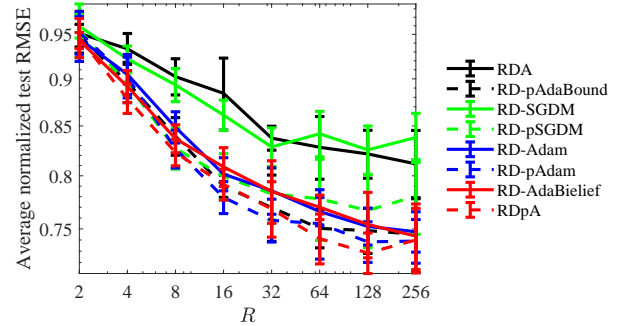}
\caption{Average normalized test RMSEs of the eight algorithms versus the number of rules $R$ on the 22 datasets.} \label{fig:6}
\end{figure}

\begin{enumerate}
\item AdaBelief performed the best among the four vanilla MBGD optimization approaches, in most cases when $R$ changed from small to large.
\item Powerball gradients always improved the performance of AdaBound, SGDM, Adam, and AdaBilief.
\item On average, Powerball AdaBelief (FCM-RDpA) achieved the best performance among the eight algorithms.
\end{enumerate}

\subsection{An Additional Approach}

This section proposes an additional approach FCM-RDpAx, to further enhance FCM-RDpA. Since in FCM-RDpA the number of rules is decoupled from the number of features, we can concatenate the original features with their (trainable) linear projections in both the antecedents and the consequents to increase the model capacity. Table~\ref{tab:DR} lists several different feature projection concatenation approaches.

\begin{table}[htpb] \centering
\caption{Comparison of four algorithms with different feature projection concatenation approaches.} \label{tab:DR}
\begin{threeparttable}
\begin{tabular}{lcc}
\toprule
Algorithm                       & Input to Antecedent            & Input to Consequent             \\ \midrule
FCM-RDpA\tnote{1}               & $\bm{X}$                       & $\bm{X}$                        \\
FCM-RDpA$'$\tnote{2}            & $[\bm{X}'\ \bm{X}]$            & $\bm{X}$                        \\
FCM-RDpA$''$\tnote{3}           & $[\bm{X}''\ \bm{X}]$           & $[\bm{X}''\ \bm{X}]$            \\
FCM-RDpAx\tnote{4}              & $[\bm{X}'\ \bm{X}]$            & $[\bm{X}'''\ \bm{X}]$           \\ \bottomrule
\end{tabular}
\begin{tablenotes}
\item[1] $\bm{X}=(\bm{x}_1,\dots,\bm{x}_N)^T\in \mathbb{R}^{N\times M}$ is the training data matrix.
\item[2] $\bm{X}'\in \mathbb{R}^{N\times \log_{2}(R)}$ is linear feature projection of $\bm{X}$, updated by the gradients of the antecedents.
\item[3] $\bm{X}''\in \mathbb{R}^{N\times \log_{2}(R)}$ is linear feature projections of $\bm{X}$, updated by the gradients of both the antecedents and the consequents.
\item[4] $\bm{X}'''\in \mathbb{R}^{N\times \log_{2}(R)}$ is linear feature projections of $\bm{X}$, updated by the gradients of the consequents.
\end{tablenotes}
\end{threeparttable}
\end{table}

\begin{figure}[htpb] \centering
\subfigure[]{\label{fig:8-R}     \includegraphics[width=.65\linewidth,clip]{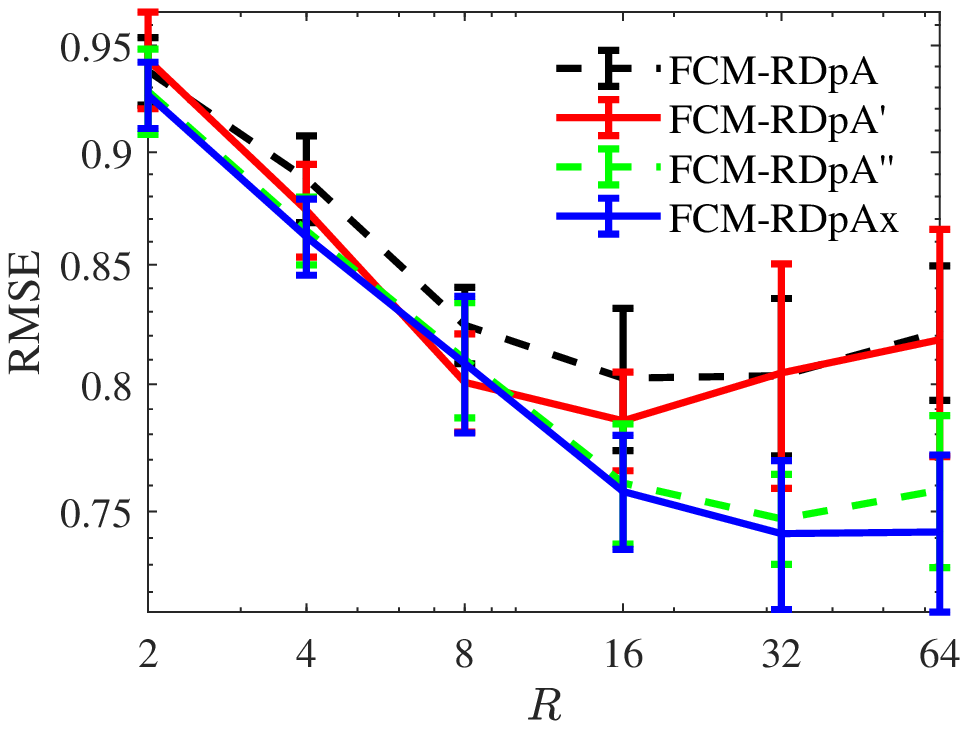}}
\subfigure[]{\label{fig:8t-R}     \includegraphics[width=.65\linewidth,clip]{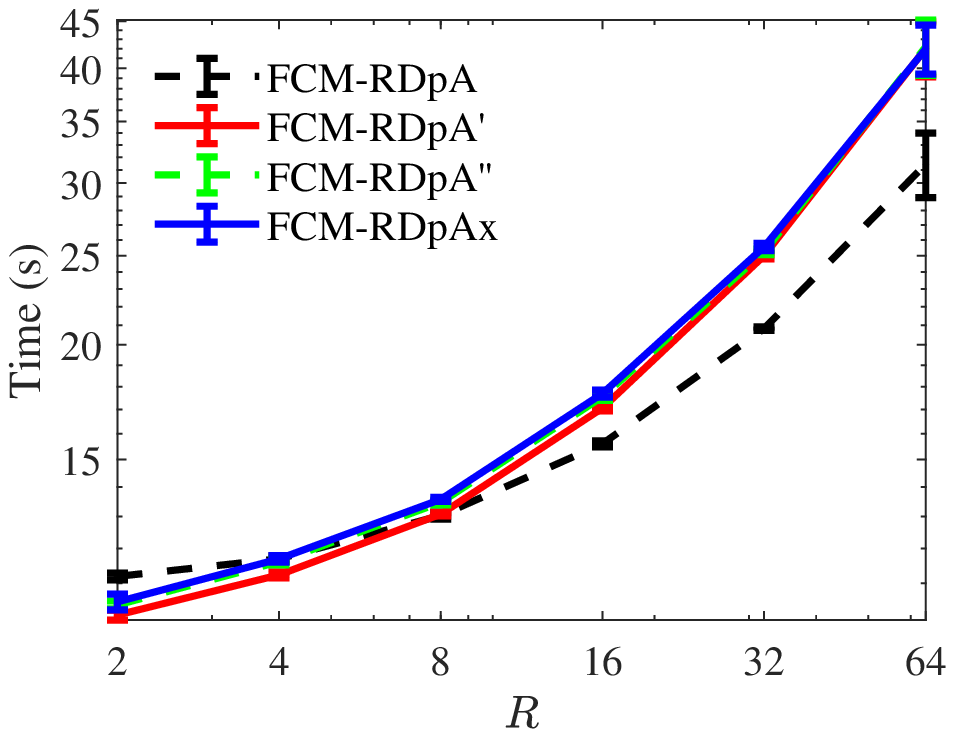}}
\caption{(a) Average normalized test RMSEs and (b) average computing time (seconds; including training, validation and test time) w.r.t. the number of rules $R$, averaged over the 22 regression datasets in eight runs with different data partitions.} \label{fig:8}
\end{figure}

The average normalized test RMSEs and the average computing time of the four algorithms (with again RR as the baseline) versus the number of rules $R$, averaged on the 22 regression datasets in eight runs with different data partitions, are shown in Fig.~\ref{fig:8}. FCM-RDpAx that utilizes the largest number of features achieved the smallest average test RMSEs with slightly higher computational cost than FCM-RDpA.

\section{Conclusions and Future Work}

This paper has proposed FCM-RDpA to enhance MBGD-RDA via fuzzy c-means clustering to initialize independent MFs, and Powerball AdaBelief to automatically adapt the gradients and learning rates, for TSK regression model optimization. Experiments on 22 regression datasets validated its effectiveness. We also proposed an additional approach, FCM-RDpAx, that can utilize more features and achieve higher generalization performance with slightly more computational cost than FCM-RDpA.

Our future research will enhance FCM-RDpA via patch learning \cite{Wu2020b}, and extend FCM-RDpA to classification and multi-view high-dimensional problems \cite{Shi2020}.


\end{document}